\documentclass[runningheads]{llncs}
\usepackage{svg}
\usepackage{graphicx}
\usepackage[export]{adjustbox}
\usepackage{enumerate}
\usepackage{amssymb}
\usepackage[subrefformat=parens,labelformat=parens]{subfig}
\usepackage{amsmath}
\usepackage{multirow}
\usepackage{booktabs}
\usepackage{textgreek}
\usepackage{hhline}
\usepackage{array}
\usepackage{tabu}
\usepackage{verbatim} 
\usepackage{graphics}
\usepackage{color}
\usepackage{url}
\usepackage[colorlinks = true]{hyperref}

\begin{document}
\title{DocSynth: A Layout Guided Approach for Controllable Document Image Synthesis}
\titlerunning{A Layout Guided Approach for Document Image Synthesis}
\author{Sanket Biswas\inst{1}\orcidID{0000-0001-6648-8270} \and
Pau Riba\inst{1}\orcidID{0000-0002-4710-0864} \and
Josep Lladós\inst{1}\orcidID{0000-0002-4533-4739} \and
Umapada Pal\inst{2}\orcidID{0000-0002-5426-2618}}
% %
\authorrunning{S.Biswas et al.}
\institute{Computer Vision Center \& Computer Science Department  \\
              Universitat Autònoma de Barcelona, Spain \\
%              Fax: +123-45-678910\\
              \email{\{sbiswas, priba, josep\}@cvc.uab.es} 
              %  \\
              \and
              CVPR Unit, Indian Statistical Institute, India \\ 
              \email{umapada@isical.ac.in}}
% %
\maketitle              % typeset the header of the contribution
\begin{abstract}
Despite significant progress on current state-of-the-art image generation models, synthesis of document images containing multiple and complex object layouts is a challenging task. This paper presents a novel approach, called DocSynth, to automatically synthesize document images based on a given layout. In this work, given a spatial layout (bounding boxes with object categories) as a reference by the user, our proposed DocSynth model learns to generate a set of realistic document images consistent with the defined layout. Also, this framework has been adapted to this work as a superior baseline model for creating synthetic document image datasets for augmenting real data during training for document layout analysis tasks. Different sets of learning objectives have been also used to improve the model performance. Quantitatively, we also compare the generated results of our model with real data using standard evaluation metrics. The results highlight that our model can successfully generate realistic and diverse document images with multiple objects. We also present a comprehensive qualitative analysis summary of the different scopes of synthetic image generation tasks. Lastly, to our knowledge this is the first work of its kind.

\end{abstract}
\section{Introduction}\label{s:intro}

The task of automatically understanding a document is one of the most significant and primary objectives in the Document Analysis and Recognition community. Nowadays, especially in business processes, paper scanned and digitally born documents coexist. There is a big variability in real-world documents coming from different domains (forms, invoices, letters, etc.). Modern Robotic Process Automation (RPO) tools in paperless offices have a compelling need for managing automatically the information of document workflows, which can integrate both reading and understanding. According to the standardized recommendations of the Office Document Architecture (ODA) \cite{horak1985office}, a document representation could be expressed by formalisms that obey two crucial aspects. The first one considers a document as an image for printing or displaying, while the second one considers its textual and graphical representation for interpreting its layout and logical structure. 

The layout structure of a document is fundamentally represented by layout objects (e.g. text or graphic blocks, images, tables, lines, words, characters and so on) while the logical structure describes the semantic relationship between conceptual elements (e.g. company logo, signature, title, body or paragraph region and so on). The recognition of document layout has been one of the most challenging problems for decades. The understanding of layout is a necessary step towards the extraction of information. Business intelligence processes require the extraction of information from document contents at large scale, for subsequent decision-making actions. Many examples can be found in different X-tech areas: fin-tech (analyze sales trends based on intelligent reading of invoices), legal-tech (determine if a clause of a contract has been violated), insurance-tech (liability from accident statement understanding). Document layout syntactically describes the whole document, and therefore allows to give context to the individual components (named entities, graphical symbols, key-value associations). Thus, performance in information extraction is boosted when it is driven by the layout. As in many other domains, the deep learning revolution has open new insights in the layout understanding problem. Consequently, there is a need for annotated data to supervise the learning tasks. Having big amounts of data is not always possible in real scenarios. In addition to the manual effort to annotate layout components, such types of images have privacy restrictions (personal data, corporate information) which prevent companies and organizations to disclose it. Data augmentation strategies are a good solution. Among the different strategies for augmenting data, synthetic generation of realistic images is one of the most successful. 

\begin{figure}[h]
    \centering
    \includegraphics[width=\linewidth]{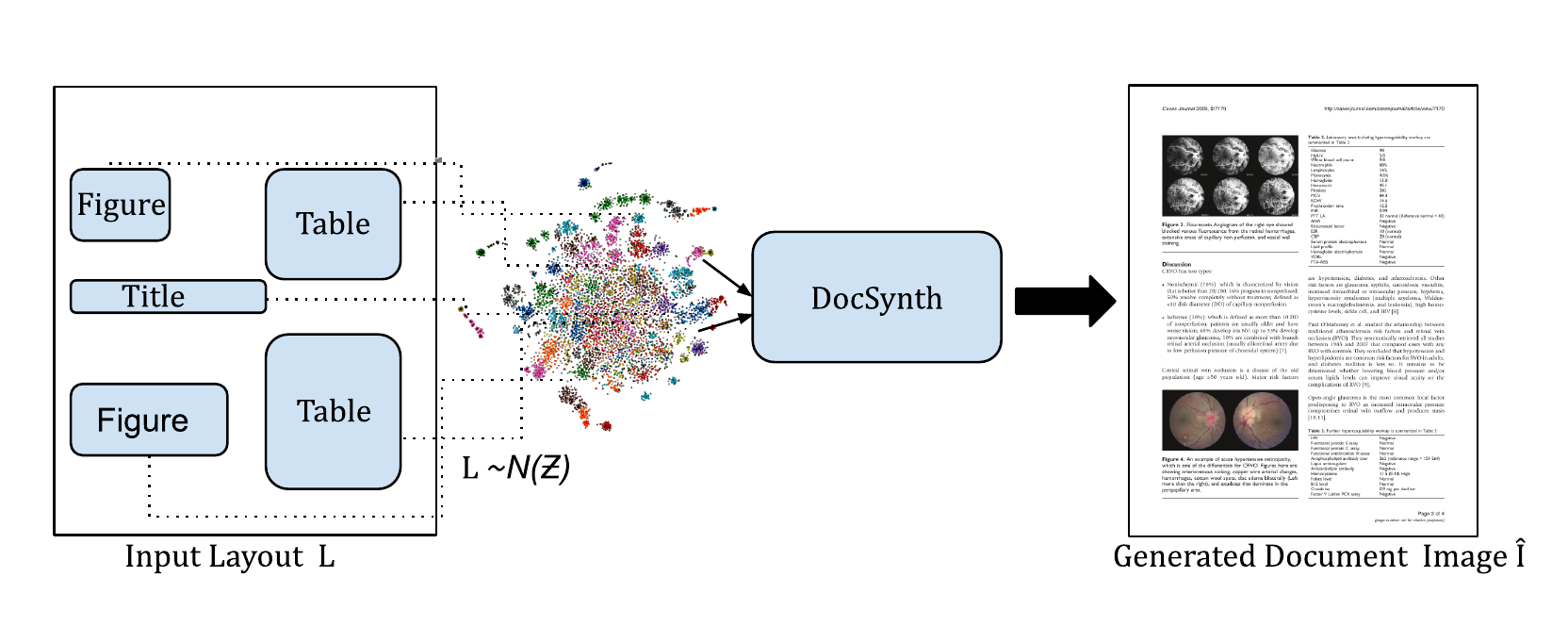}
    % \includesvg[width=\linewidth]{figures/Teaser_ICDAR_2021.svg}
    % \captionsetup{justification=centering,margin=1cm}
    \caption{\textbf{Illustration of the Task: }Given an input document layout with object bounding boxes and categories configured in an image lattice, our model samples the semantic and spatial attributes of every layout object from a normal distribution, and generate multiple plausible document images as required by the user.}
    \label{fig:teaser_icdar}
\end{figure}

This work discusses a research effort which is intended to develop a synthetic generation tool called DocSynth for rendering realistic printed documents with plausible layout objects desired by the user. A simple illustration of this task is as shown in Figure \ref{fig:teaser_icdar}.
The proposed model is able to generate samples given a single reference layout image.
Thus, it can generate training data with one single sample per class and can be adapted to few-shot settings for document classification tasks. 
This automatic document image synthesizer could provide a possible solution to manage all papers as well as electronic documents with a centralized platform manipulated by the user. In fact, this practical application has the potential to improve visual search and information retrieval engines.  Usually, in retrieval task the user wants to index in a repository of documents (real ones). Instead one could create variations of the query sample to improve retrieval performances of the model.

While  classical  computer  graphics techniques have been used in modeling for example geometry, projections, surface properties and cameras, the more recent computer vision techniques rely on the quality of designed machine learning approaches to learn from real world examples to generate synthetic images. 
Explicit reconstruction and rendering of document properties (both graphical and textual) in the form of complex layout objects is a hard task from both computer graphics and vision perspective. To this end, traditional image-based rendering approaches tried to overcome these issues, by using simple heuristics to combine a captured imagery. But applying these heuristical approaches for synthesizing images with complex document layouts generate artifacts and does not provide an optimal solution. Neural rendering brings the promise of addressing the problem of both reconstruction and rendering by using deep generative models like Generative Adversarial Networks (GANs) and Variational Auto encoders (VAEs) and to learn  complex mappings  from  captured  images  to  novel  images. They help to combine physical knowledge, e.g., mathematical  models  of  projection and geometry,  with  learned  components  to  yield  new and powerful algorithms for controllable image generation.

The main contributions of this work are as follows.

\begin{enumerate}
    \item A new model is proposed for synthetic document image generation  guided by the layout of a reference sample.
    \item Qualitative and quantitative results on the PubLayNet dataset \cite{zhong2019publaynet}, demonstrate our model's capability to generate complex layout documents with respect to spatial and semantic information of object categories.
    \item Also this work addresses the layout-guided document image synthesis task with an analytical understanding as the first of its kind in the document analysis community. 
\end{enumerate}

The rest of this paper is organized as follows: in Section \ref{s:soa} we review the relevant literature. Section \ref{s:method} describes the main methodological contribution of the work. In Section \ref{s:results} we provide a quantitative and qualitative experimental analysis. Finally, Section \ref{s:conclusion} draws the main contributions and outlines future perspectives.

\section{Related Work}\label{s:soa}
The analysis of structural and spatial relations between complex layouts in documents has been a significant challenge in the field of Document Analysis and Recognition. Extracting the physical and logical layout in documents is a required step in tasks such as Optical Character Recognition for document image transcription, document classification, or information extraction. The reader is referred to \cite{binmakhashen2019document} for a comprehensive survey on the state of the art on document layout analysis. 

As it has been introduced in section \ref{s:intro} the main objective of this work is to construct a generative neural model to construct visually plausible document images given a reference layout. The strategy for augmenting data and its corresponding ground truth by synthetic images automatically generated has gained interest among the Computer Vision community. Since they were proposed by Goodfellow in 2014, Generative Adversarial Networks (GANs) \cite{goodfellow_generative_2014} and subsequent variants have been a successful method to generate realistic images, ranging from handwritten digits to faces and natural scenes. A step forward which is a scientific challenge in the controlled generation of images in terms of the composition of objects and their arrangement.
Lake et al. in \cite{lake2015human} suggested a hierarchical generative model that can build whole objects from individual parts, it is shown to generate Omniglot characters as a composition of the strokes. Zhao et al. \cite{Zhao2020Layout2Image} proposed a model that can generate a set of realistic images with objects in the desired locations, given a reference spatial distribution of bounding boxes and object labels. Our work has been inspired in this work.
 
Preserving the reliable representation of layouts has shown to be very useful in various graphical design contexts, which typically involve highly structured and content-rich objects. One such recent intuitive understanding was established by Li et al. \cite{li2019layoutgan} in their LayoutGAN, which aims to generate realistic document layouts using Generative Adversarial Networks (GANs) with a wireframe rendering layer. Zheng et. al. \cite{zheng2019content} used a GAN-based approach to generate document layouts but their work focused mainly on content aware generation, that primarily uses the content of the document as an additional prior. To use more highly structured object generation, it is very important to focus operate on the low dimensional vectors unlike CNN's. Hence, in the most recent literature, Patil et. al. \cite{patil2020read} has come up with a solution called 'READ' that can make use of this highly structured positional information along with content to generate document layouts. Their recursive neural network-based resulting model architecture provided state-of-the-art results for generating synthetic layouts for 2D documents. But their solution could not be applicable for document image level analysis problems. Kang et. al. \cite{kang2020ganwriting} actually exploited the idea to generate synthetic data at the image level for handwritten word images. 

Summarizing, state-of-the-art generative models are still unable to produce plausible yet diverse images for whole page documents. In this work we propose a direction to condition a generative model for whole page document images with synthesized variable layouts. 
% Figure \ref{fig:layout_extraction} shows an example of how they make use of relational information between the layouts on an example document. 

\section{Method}\label{s:method}
In this section we describe the contributions of the work. We first formally formulate the problem, and introduce the basic notation. Afterwards, we describe the proposed approach, the network structure, its learning objectives, and finally the implementation details.

% \begin{figure}[h]
%     \centering
%     % \includegraphics[width=\linewidth]{figures/Teaser_ICDAR_2021.svg}
%     \includesvg[width=\linewidth]{figures/Model_Diagram.svg}
%     % \captionsetup{justification=centering,margin=1cm}
%     \caption{\textbf{Illustration of the Task: }Given an input document layout with object bounding boxes and categories configured in an image lattice, our model samples the semantic and spatial attributes of every layout object from a normal distribution, and generate multiple plausible document images as required by the user.}
%     \label{fig:model_icdar}
% \end{figure}

\begin{figure}[h]
    \centering
    \includegraphics[width=\linewidth]{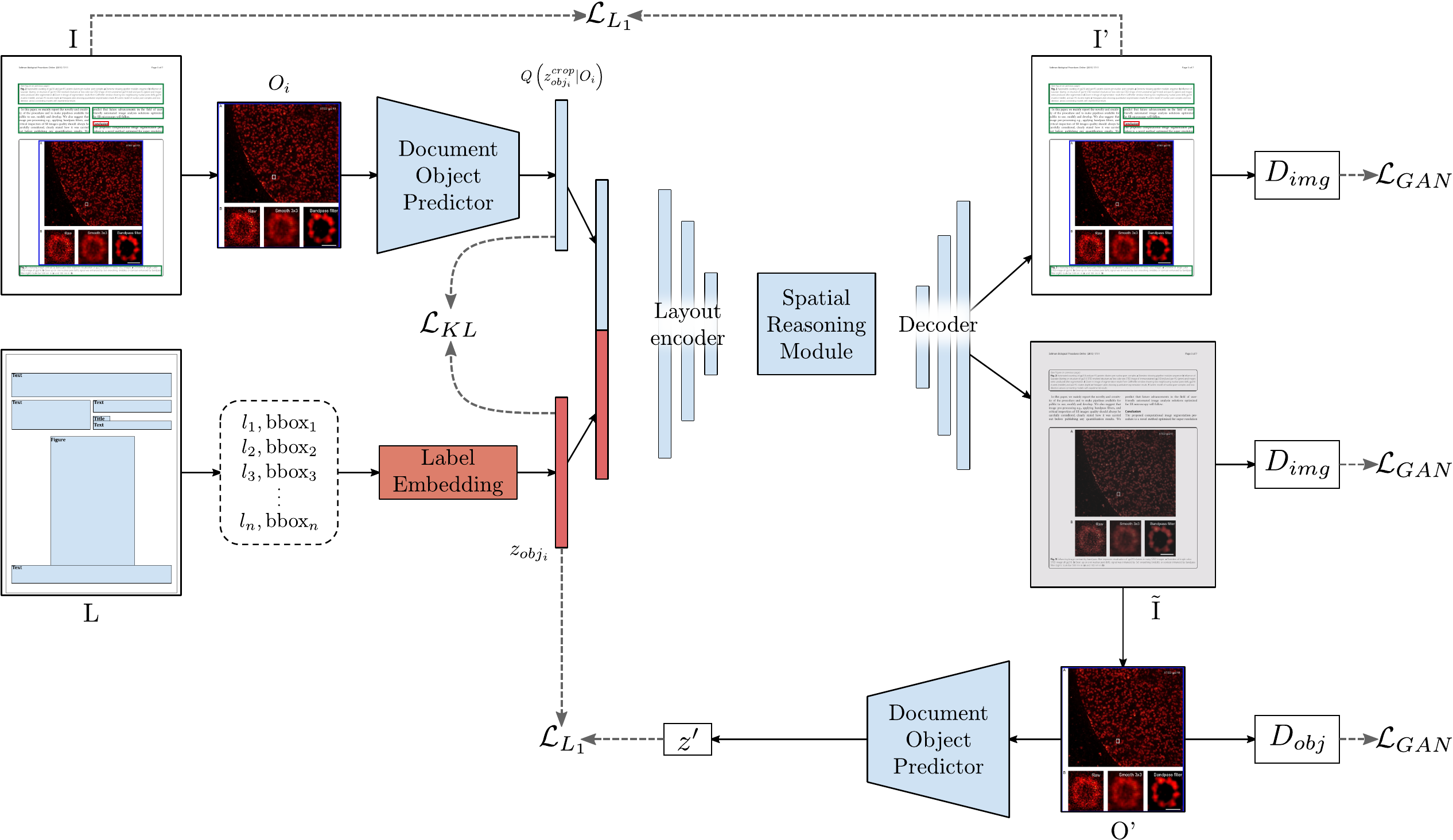}
    % \captionsetup{justification=centering,margin=1cm}
    \caption{\textbf{Overview of our DocSynth Framework:}The model has been trained adversarially against a pair of discriminators and a set of learning objectives as depicted.}
    \label{fig:model_icdar}
\end{figure}

\subsection{Problem Formulation}

Let us start by defining the problem formally. Let $X$ be an image lattice (e.g. of size 128x128) and $I$ be a document image defined on the lattice. Let $L=\left\{\left(\ell_{i}, \text { bbox }_{i}\right)_{i=1}^{n}\right\}$ be a layout which contains $n$ labeled object instances with defined class categories $\ell_{i} \in {O}$ and bounding boxes of these instances represented by top-left and bottom-right coordinates on the canvas, $\text { bbox }_{i} \subset X$, and $|{O}|$ is the total number of document object categories (eg. table, figure, title and so on). Let $Z_{o b j}$ be the overall sampled latent estimation comprising every object instance ${O}_{i}$ in the layout $L$, which can be represented as $Z_{o b j}=\left\{\mathbf{z}_{o b j_{i}}\right\}_{i=1}^{n}$. The latent estimation have been sampled randomly for the objects from the standard prior Normal distribution $\mathcal{N}(0,1)$ under the i.i.d. setup.                  

The layout-guided document image synthesis task can be codified as learning a generator function G which can map a given document layout input $(L, Z_{o b j})$ to the generated output image $\tilde{I}$ as shown in eqn. \ref{eq:gen_function}. 

\begin{equation}
    \tilde{I}=G\left(L, Z_{o b j} ; \Theta_{G}\right)
\label{eq:gen_function}
\end{equation}
where $\Theta_{G}$ represents the parameters of the generation function $G$ which needs to be learned by our model. Primarily, a generator model $G(.)$ is able to capture the underlying conditional data distribution $p(\tilde{I}|L, Z_{o b j} ; \Theta_{G})$ present in a higher dimensional space, equivariant with respect to spatial locations of document layout objects ${\text{bbox}}_{i}$. 

The proposed model in this work for the above formulated task investigates three different challenges: (1) Given the user provides the input document layout $L$, is the model capable of synthesizing plausible document images while preserving the object properties conditioned on $L$? (2) Given the user provides the input document layout $L$, can the model generate multiple variable documents using different style $\mathbf{z}_{o b j_{i}}$ of objects while retaining the object configuration $\ell_{i}, \text { bbox }_{i}$ in the input layout? (3) Given a tuple $(L, Z_{o b j})$ , is the generator capable of generating consistent document images for different $(\tilde{L}, \tilde{Z_{o b j}})$ where a user can add an object to existing layout $L$ or just modify the location or label of existing objects?

Handling such complexities using deep generative networks is difficult due to the difficulty of sampling the posterior elements. This work focuses to tackle the problem by designing a single generator model $G(\cdot)$ that tries to provide an answer to the above mentioned research questions. 

\subsection{Approach}

% In this section we present the technical details of our proposed model for the task. 
To build on the layout-guided synthetic document generation pipeline, we aim to explain our proposed approach in two different parts: Training and Inference.

\subsubsection{Training:}

The overall training pipeline of our proposed approach is illustrated in Figure \ref{fig:model_icdar}. Given an input document image $I$ and its layout $L=\left\{\left(\ell_{i}, \text { bbox }_{i}\right)_{i=1}^{n}\right\}$, the proposed model creates a category label embedding $e_{i}$ for every object instance $O_{i}$ in the document. A set of object latent estimations $Z_{o b j}=\left\{\mathbf{z}_{o b j_{i}}\right\}_{i=1}^{n}$ are sampled from the standard prior normal distribution $\mathcal{N}(0,1)$, while another set of object latent estimations $Z_{o b j}^{crop} = \left\{\mathbf{z}_{o b j i} ^{crop}\right\}_{i=1}^{n}$  are sampled from the posterior distribution $Q\left(\mathbf{z}_{o b j i}^{crop} \mid O_{i}\right)$ conditioned on the features received from the cropped objects $O_{i}$ of input image $I$ as shown in Figure \ref{fig:model_icdar} in the document object predictor. This eventually allows us to synthesize two different datasets: 
(1) A collection of reconstructed images $I'$ from ground-truth image $I$ during training by mapping the input $(L, Z_{o b j}^{crop})$ through the generator function $G$.
(2) A collection of generated document images $\tilde{I}$ by mapping $(L, Z_{o b j})$ through $G$, where the generated images match the original layout but exhibits variability in object instances which is sampled from random distribution.
To allow consistent mapping between the generated object $\tilde{O}$ in $\tilde{I}$ and sampled $Z_{o b j}$ , the latent estimation is regressed by the document object predictor as shown in Figure \ref{fig:model_icdar}. The training is done with an adversarial approach by including two discriminators to classify the generated results as real or fake at both image-level and object-level.     

\subsubsection{Inference:}

During inference time, the proposed model synthesizes plausible document images from the layout $L$ provided by the user as input and the object latent estimation $Z_{o b j}$ sampled from the prior $\mathcal{N}(0,1)$ as illustrated in the Figure \ref{fig:teaser_icdar}. 

\subsection{Generative Network}

The proposed synthetic document image generation architecture consists of mainly three major components: two object predictors $E$ and $E'$, a conditioned image generator $H$, a global layout encoder $C$, an image decoder $K$ and an object and image discriminator denoted by $D_{obj}$ and $D_{img}$ respectively.

\subsubsection{Object Encoding:} 

Object latent estimations $Z_{o b j}^{crop}$ are first sampled from the ground-truth image $I$ with the object predictor $E$. They help to model variability in object appearances, and also to generate the reconstructed image $I'$. The object predictor $E$ predicts the mean and variance of the posterior distribution for every cropped object $O_{i}$ from the input image. 
To boost the consistency between the generated output image $\tilde{I}$  and its object estimations, the model also has another predictor $E'$ which infers the mean and variances for the generated objects $O'$ cropped from $\tilde{I}$. The predictors $E$ and $E'$ consist of multiple convolutional layers with two dense fully-connected layers at the end.

\subsubsection{Layout Encoding:}

Once the object latent estimation $\mathbf{z}_{i} \in \mathbb{R}^{n}$ has been sampled from the posterior or the prior distribution $\left(\mathbf{z}_{i} \in\left\{Z_{o b j}^{crop}, Z_{o b j}\right\}\right)$, the next step is to construct a layout encoding denoted by $F_{i}$ with the input layout information $L=\left\{\left(\ell_{i}, \text { bbox }_{i}\right)_{i=1}^{n}\right\}$ as provided by the user for every object $O_{i}$ in the image $I$. Each feature map $F_{i}$ should contain the disentangled spatial and semantic information corresponding to layout $L$ and appearance of the objects $O_{i}$ interpolated by latent estimation $\mathbf{z}_{i}$. The object category label $\ell_{i}$ is transformed as a label embedding $e_{i} \in \mathbb{R}^{n}$ and then concatenated with the latent vector $\mathbf{z}_{i}$. The resultant feature map $F_{i}$ for every object is then filled with the corresponding bounding box information $\text {bbox}_{i}$ to form a tuple represented by $<\ell_{i}, \mathbf{z}_{i}, \text{bbox}_{i}>$. These feature maps encoding this layout information are then fed to a global layout encoder network $C$ containing multiple convolutional layers to get downsampled feature maps.                        

\subsubsection{Spatial Reasoning Module:}

Since the final goal of the model is to generate plausible synthetic document images with the encoded input layout information, the next step of the conditioned generator $H$ would be to generate a good hidden feature map $h$ to fulfill this objective. The hidden feature map $h$ should be able to perform the following: (1) encode global features that correlate an object representation with its neighbouring ones in the document layout (2) encode local features with spatial information corresponding to every object (3) should invoke spatial reasoning about the plausibility of the generated document with respect to its contained objects.  

To meet these objectives, we choose to define the spatial reasoning module with a convolutional Long-Short-Term Memory(conv-LSTM) network backbone. Contrary to vanilla LSTMs,conv-LSTMs replace the hidden state vectors with feature maps instead. The different gates in this network are also encoded by convolutional layers, which also helps to preserve the spatial information of the contents more accurately. The conv-LSTM encodes all the object feature maps $F_{i}$ in a sequence-to-sequence manner, until the final output of the network gives a hidden layout feature map $h$. 

\subsubsection{Image Reconstruction and Generation:}

Given the hidden layout feature map $h$ already generated by the spatial reasoning module, we move towards our final goal for the task. An image decoder $K$ with a stack of deconvolutional layers is used to decode this feature map $h$ to two different images, $I'$ and $\tilde(I)$. The image $I'$ is reconstructed from the input image $I$ using latent estimation $Z_{o b j}^{crop}$ conditioned on its objects $O$. The image $\tilde(I)$ is the randomly generated image using $Z_{o b j}$ directly sampled from the prior $\mathcal{N}(0,1)$. Both these images retain the same layout structure as mentioned in the input. 

\subsubsection{Discriminators:}

To make the synthetic document images look realistic and its objects noticeable, a pair of discriminators $D_{img}$ and $D_{obj}$ is adopted to classify an input image as either real or fake by maximizing the GAN objective as shown in eqn. \ref{eq:gan}. But, the generator network $H$ is being trained to minimize $\mathcal{L}_{\mathrm{GAN}}$. 

\begin{equation}
    \mathcal{L}_{\mathrm{GAN}}=\underset{x \sim p_{\mathrm{rcal}}}{\mathbb{E}} \log D(x)+\underset{y \sim p_{\mathrm{fake}}}{\mathbb{E}} \log (1-D(y))
\label{eq:gan}
\end{equation}

While the image discriminator $D_{img}$ is applied to input images $I$, reconstructed images $I'$ and generated sampled images  $\tilde{I}$, the object discriminator $D_{obj}$ is applied at the object-level to assess the quality of generated objects $O'$ and make them more realistic.

\subsection{Learning Objectives}

The proposed model has been trained end-to-end in an adversarial manner with the generator framework and a pair of discriminators. The generator framework, with all its components help to minimize the different learning objectives during training phase. Our GAN model makes use of two adversarial losses: image adversarial loss $\mathcal{L}_{\mathrm{GAN}}^{\mathrm{img}}$ and object adversarial loss $\mathcal{L}_{\mathrm{GAN}}^{\mathrm{obj}}$. Four more losses have been added to our model, including KL divergence loss $\mathcal{L}_{\mathrm{KL}}$, image reconstruction loss $\mathcal{L}_{1}^{\mathrm{img}}$, object reconstruction loss $\mathcal{L}_{1}^{\mathrm{obj}}$ and auxiliary classification loss $\mathcal{L}_{\mathrm{AC}}^{\mathrm{obj}}$, to enhance our synthetic document generation network.

% \begin{equation}
%     \mathcal{L}_{\mathrm{KL}}=\sum_{i=1}^{o} \mathbb{E}\left[\mathcal{D}_{\mathrm{KL}}\left(Q\left(\mathbf{z}_{r i} \mid \mathbf{O}_{i}\right) \| \mathcal{N}\left(\mathbf{z}_{r}\right)\right)\right]
% \label{eq:KL}
% \end{equation}
The overall loss function used in our proposed model can be defined as shown in equation \ref{eq:overall}:
\begin{equation}
    \mathcal{L}_{G}=\lambda_{1} \mathcal{L}_{\mathrm{GAN}}^{\mathrm{img}}+\lambda_{2} \mathcal{L}_{\mathrm{GAN}}^{\mathrm{obj}}+\lambda_{3} \mathcal{L}_{\mathrm{AC}}^{\mathrm{obj}}+\lambda_{4} \mathcal{L}_{\mathrm{KL}}+\lambda_{5} \mathcal{L}_{1}^{\mathrm{img}}+\lambda_{6} \mathcal{L}_{1}^{\text {obj}}
\label{eq:overall}
\end{equation}

\subsection{Implementation details}

In order to stabilise training for our generative network, we used the Spectral-Normalization GAN \cite{miyato2018spectral} as our model backbone. We used conditional batch normalization \cite{de2017modulating} in the object predictors to better normalize the object feature maps. The model has been adapted for 64x64 and 128x128 image sizes. The values of the six hyperparameters $\lambda_{1}$ to $\lambda_{6}$ are set to 0.01, 1, 8, 1,1 and 1, respectively. These values have been set experimentally. The Adam optimizer \cite{kingma2014adam} was to train all the models with batch size of 16 and 300,000 iterations in total. For more finer details, we will make our code publicly available.

\section{Experimental Validation}\label{s:results}
\begin{figure}[h]
    \centering
    \includegraphics[width=\linewidth]{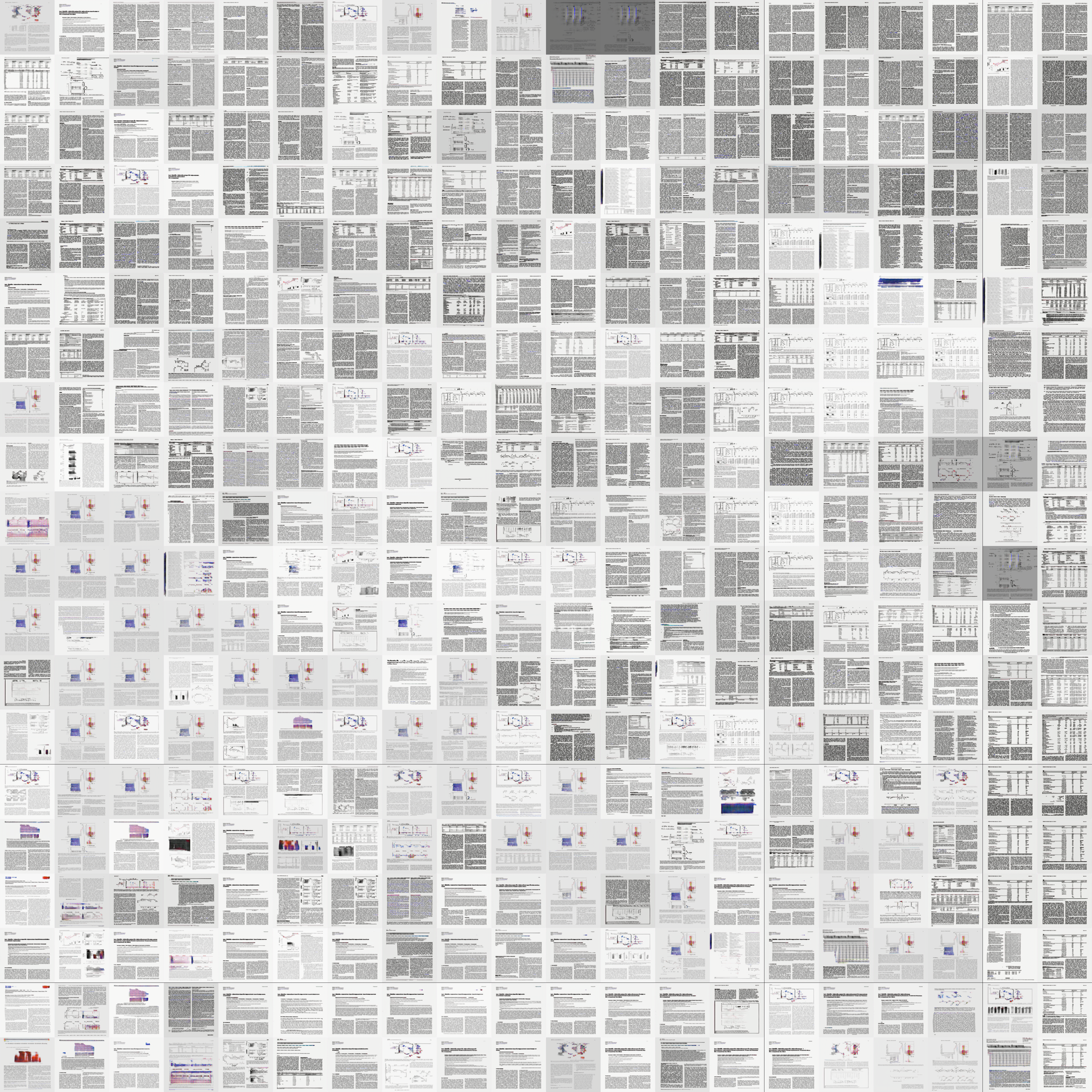}
    % \includesvg[width=\linewidth]{figures/tsne_grid_synthetic.png}
    % \captionsetup{justification=centering,margin=1cm}
    \caption{t-SNE visualization of the generated synthetic document images}
    \label{fig:results_tsne}
\end{figure}

\begin{figure}[h]
    \centering
    \includegraphics[width=\linewidth]{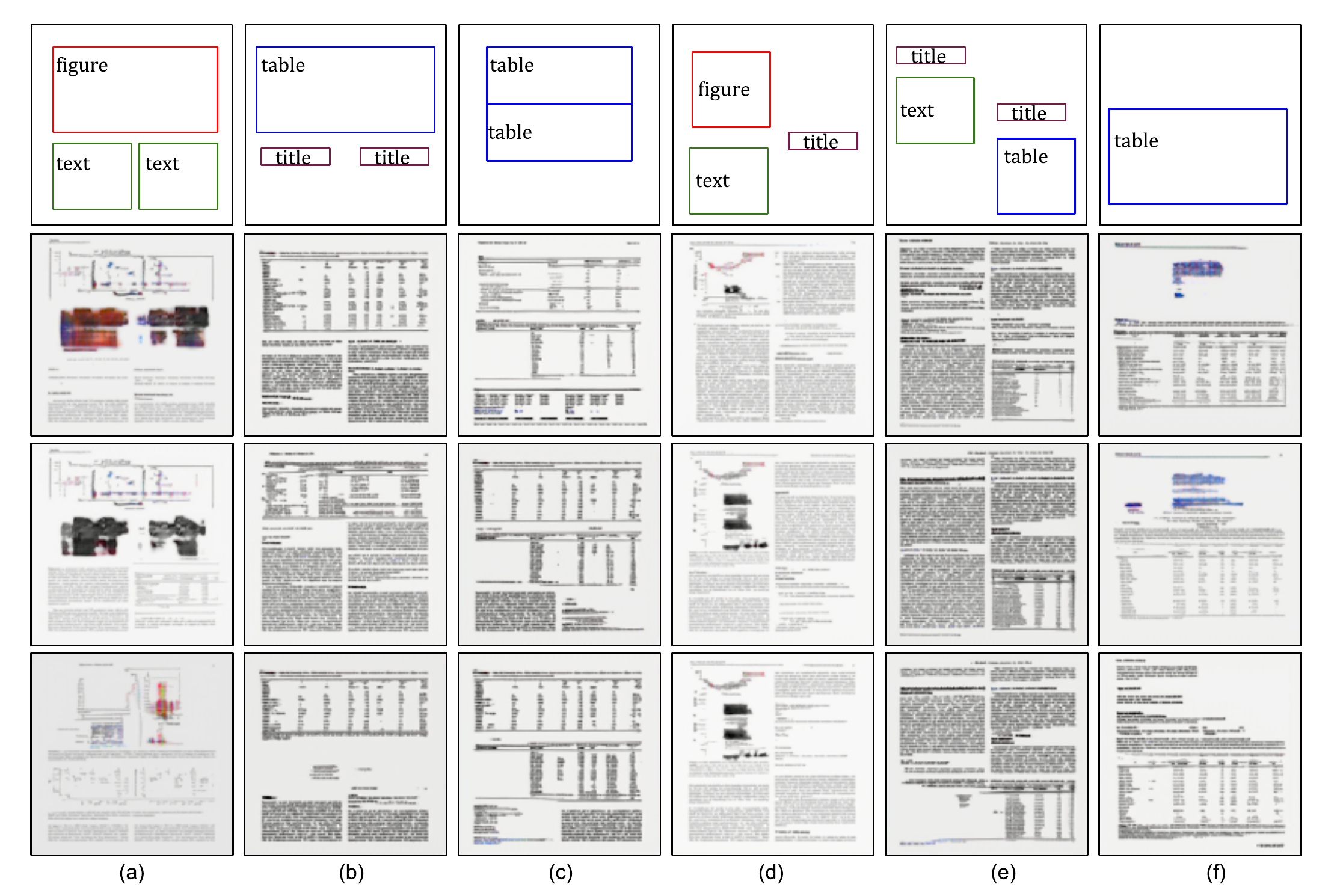}
    % \includesvg[width=\linewidth]{figures/results_diversity.pdf}
    % \captionsetup{justification=centering,margin=1cm}
    \caption{\textbf{Examples of diverse synthesized documents generated from the same layout: }Given an input document layout with object bounding boxes and categories, our model samples 3 images sharing the same layout structure, but different in style and appearance.}
    \label{fig:results_diversity}
\end{figure}

\begin{figure}[h]
    \centering
    \includegraphics[width=\linewidth]{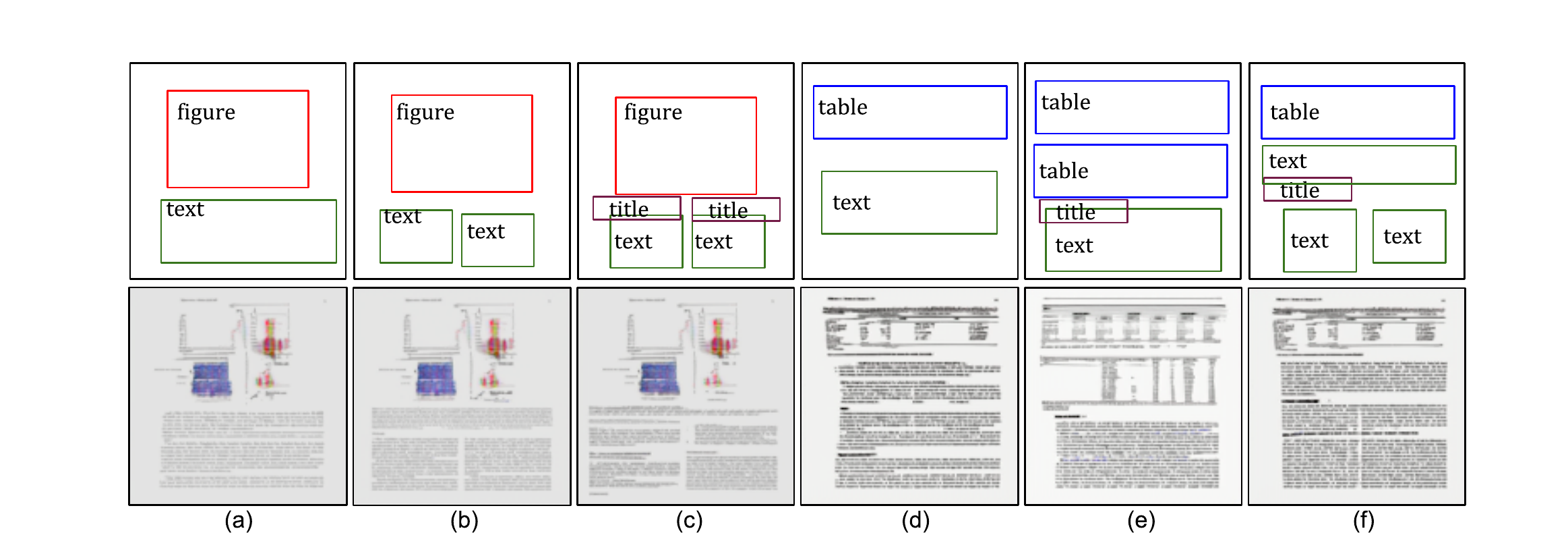}
    % \includesvg[width=\linewidth]{figures/results_add_bbox_ref.svg}
    % \captionsetup{justification=centering,margin=1cm}
    \caption{\textbf{Examples of synthesized document images by adding or removing bounding boxes based on previous layout:}There are 2 groups of images (a)-(c) and (d)-(f) in the order of adding or removing objects.}
    \label{fig:results_add_bbox}
\end{figure}

Extensive experimentation was conducted to evaluate our adapted DocSynth framework. Since this work introduces the first fundamental approach towards the problem of layout-guided document image synthesis, we try to conduct some ablation studies that are important for proposing our model as a superior baseline for the task. Also, we try to analyse our obtained results with the model, both qualitatively and quantitatively. All the code necessary to reproduce the experiments is available at \href{https://github.com/biswassanket/synth_doc_generation}{github.com/biswassanket/synth\_doc\_generation} using the PyTorch framework.

\subsection{Datasets}

We evaluate our proposed DocSynth framework on the PubLayNet dataset~\cite{zhong2019publaynet} which mainly contains images taken from the PubMed Central library for scientific literature. There are five defined set of document objects present in this dataset: text, title, lists, tables and figures. The entire dataset comprises 335,703 images for training and 11,245 images for validation.

\subsection{Evaluation Metrics}

Plausible document images generated from layout should fulfill the following conditions: (1) They should be realistic (2) They should be recognizable (3) They should have diversity. In this work, we have adapted two different evaluation metrics for evaluating our rendered images for the problem.

\subsubsection{Fréchet Inception Distance (FID):}

The FID metric ~\cite{heusel2017gans} is a standard GAN performance metric to compute distances between the feature vectors of real images and the feature vectors of synthetically generated ones. A lower FID score denotes a better quality of generated samples and more similar to the real ones. In this work, the Inception-v3 ~\cite{szegedy2016rethinking} pre-trained model were used to extract the feature vectors of our real and generated samples of document images.

\subsubsection{Diversity Score:}

Diversity score calculates the perceptual similarity between two images in a common feature space. Different from FID, it measures the difference of an image pair generated from the same input. The LPIPS \cite{zhang2018unreasonable} metric actually used the AlexNet \cite{krizhevsky2012imagenet} framework to calculate this diversity score. In this work, we adapted this perceptual metric for calculating diversity of our synthesized document images.

\subsection{Qualitative Results}

\subsubsection{Generating Synthetic Document Images:}

In order to highlight the ability of our proposed model to generate diverse realistic set of document images, we present in Figure \ref{fig:results_tsne} a t-SNE ~\cite{van2008visualizing} visualization of the different synthetic data samples with plausible layout content and variability in overall style and structure. Different clusters of samples correspond to particular layout structure as observed in the figure. We observe that synthetic document samples with complex layout structures have been generated by our model. From these examples, it is also observed that our model is powerful enough to generate complex document samples with multiple objects and multiple instances of the same object category. All the generated samples from the model shown in Figure \ref{fig:results_tsne} have a dimension of 128x128. In this work, we propose two final model baselines, for generated images of 64x64 and 128x128 dimension.               

\subsubsection{Controllable Document Synthesis:}

One of the most intriguing challenges in our problem study was the controllable synthesis of document images guided by user specified layout provided as an input to our DocSynth generative network. We proposed a qualitative analysis of the challenge in two different case studies. 

In the first case study as shown in Figure \ref{fig:results_diversity}, it shows a diverse set of generated documents from a single reference layout as specified by the user. In real life scenario, documents do have the property to exhibit variability in appearance and style while preserving the layout structure. The generated samples in this case obey the spatial constraints of the input bounding boxes, and also the generated objects exhibit consistent behaviour with the input labels. 

In the second case study as shown in Figure \ref{fig:results_add_bbox}, we demonstrate our model's ability to generate documents with complex layouts by starting with a very simple layout and then adding a new bounding box or removing an existing bounding box from the input reference layout. From these results we can clearly infer that new objects can be introduced in the images at the desired locations by the user, and existing objects can be modified as new content is added.

\subsection{Quantitative Results}

Table \ref{tab:results_summary} summarizes the comparison results of the FID and Diversity Score of our proposed model baselines for both \(128\times 128\) and \(64 \times 64\) generated documents. For proper comparison, we have compared the performance scores of the model generated images with the real images. The model generated images are quite realistic as depicted by the performance scores for both FID and Diversity scores. The performance scores obtained for \(64\times 64\) image generation model are slightly better compared to those obtained for \(128\times 128\) images.           
% \usepackage{booktabs}
% \usepackage{graphicx}
\begin{table}[ht]
\centering
    \caption{Summary of the final proposed model baseline for synthetic document generation}
{%
\begin{tabular}{lcc}
\toprule
Method                                      & FID                        & Diversity Score            \\ \midrule
Real Images (\(128\times 128\)) & 30.23 & 0.125 \\ 
DocSynth (\(128\times 128\)) & \textbf{33.75} & \textbf{0.197} \\
Real Images (\(64\times 64\))   & 25.23 & 0.115 \\ 
DocSynth (\(64\times 64\))                  & \textbf{28.35}             & \textbf{0.201}             \\ \bottomrule
\end{tabular}%
}
\label{tab:results_summary}
\end{table}

\subsection{Ablation studies}

We demonstrate the importance of the key components in our model by creating some ablated models trained on the PubLayNet dataset \cite{zhong2019publaynet}. The following studies clearly illustrate the importance of these elements for solving the task.  

\subsubsection{Spatial Reasoning module:}

As already discussed, the spatial reasoning module comprising conv-LSTM to generate the hidden feature map $h$ is one of the most significant components in our model. For generating novel realistic synthetic data, we compare our model results with conv LSTM over vanilla LSTM and also modifying its number of layers for exhaustive analysis.

\begin{table}[ht]
\centering
    \caption{Ablation Study based on different Spatial Reasoning backbones used in our model}
% \resizebox{\textwidth}{!}
{%
\begin{tabular}{lc}
\toprule
Reasoning Backbone & FID   \\ \midrule
No LSTM            & 70.61 \\
Vanilla LSTM       & 75.71 \\
conv-LSTM(k=1)     & 37.69 \\
conv-LSTM(k=2)     & 36.42 \\
conv-LSTM(k=3)     & \textbf{33.75} \\ \bottomrule
\end{tabular}%
}

\label{tab:ablated_1}
\end{table}

\section{Conclusion}\label{s:conclusion}
In this work, we have presented a novel approach to automatically synthesize document images according to a given layout. The proposed method, is able to understand the complex interactions among the different layout components to generate synthetic document images that fulfill the given layout. Despite the low resolution of the generated images, we believe that this work supposes the first step towards the generation of whole synthetic documents whose contents are related to the context of the page. Indeed, other applications arise from this synthetic generation besides generating realistic images which opens a large variety of future research lines.

The future scope will be mainly focused on two research  lines. Firstly, high resolution documents with understandable content is the final goal for any synthetic document generator, therefore, we plan to extend our model towards this end. Secondly, exploiting the generated data for supervision purposes, can improve the performance on tasks such as document classification, table detection or layout analysis.

\section*{Acknowledgment}

This work has been partially supported by the Spanish projects RTI2018-095645-B-C21, and FCT-19-15244, and the Catalan projects 2017-SGR-1783, the CERCA Program / Generalitat de Catalunya and PhD Scholarship from AGAUR (2021FIB-10010).

\bibliographystyle{splncs04}
\bibliography{main}




% LNCS Homepage, \url{http://www.springer.com/lncs}. Last accessed 4
% \end{thebibliography}
\end{document}